\documentclass[10pt,twocolumn,letterpaper]{article}

\usepackage[review]{cvpr}      % To produce the REVIEW version
\usepackage{mathtools}
\usepackage{graphicx}
\usepackage{enumitem}
\usepackage{fancyhdr}
\usepackage{ragged2e}
\usepackage{subcaption}
\usepackage{amsmath}
\usepackage{amssymb}
\usepackage{amsthm}
\theoremstyle{plain}
\newtheorem{theorem}{Theorem}[section]

\theoremstyle{definition}

\theoremstyle{remark}

\usepackage{float} % For [H] placement

\title{GCSAM: Gradient Centralized Sharpness Aware Minimization}

\author{\textbf{Mohamed Hassan, Aleksandar Vakanski, Boyu Zhang, Min Xian}\\ \small{Department of Computer Science, University of Idaho}}
\date{}
\begin{document}
\maketitle
\begin{abstract}
The generalization performance of deep neural networks (DNNs) is a critical factor in achieving robust model behavior on unseen data. Recent studies have highlighted the importance of sharpness-based measures in promoting generalization by encouraging convergence to flatter minima. Among these approaches, Sharpness-Aware Minimization (SAM) has emerged as an effective optimization technique for reducing the sharpness of the loss landscape, thereby improving generalization. However, SAM's computational overhead and sensitivity to noisy gradients limit its scalability and efficiency. To address these challenges, we propose Gradient-Centralized Sharpness-Aware Minimization (GCSAM), which incorporates Gradient Centralization (GC) to stabilize gradients and accelerate convergence. GCSAM normalizes gradients before the ascent step, reducing noise and variance, and improving stability during training. Our evaluations indicate that GCSAM consistently outperforms SAM and the Adam optimizer in terms of generalization and computational efficiency. These findings demonstrate GCSAM’s effectiveness across diverse domains, including general and medical imaging tasks. Our code is available at \url{https://github.com/mhassann22/GCSAM}.
\end{abstract}    
\section{Introduction}
\label{sec:intro}

Understanding the generalization behavior of overparameterized deep neural networks (DNNs) has recently become a important area of study, as it offers valuable insights into handling overfitting and enhancing performance on unseen data ~\cite{Author1,Author2}. As the scale of models and datasets continues to expand, the development of optimization algorithms that enhance generalization capabilities becomes increasingly critical. To better understand the generalization behavior of DNNs, comprehensive empirical studies by Jiang \etal~\cite{Author3} and Dziugaite \etal~\cite{Author4} have evaluated various generalization metrics, finding that sharpness-based measures exhibit the strongest correlation with generalization performance. Based on these findings, optimizers that lead to flatter minima, rather than sharp minima, are particularly effective in promoting generalization, as flatter minima have been shown to correlate strongly with improved model performance in overparameterized setting ~\cite{Author5,Author6}.

Sharpness-Aware Minimization (SAM) ~\cite{Author7} is a promising optimization technique for finding flatter minima to improve generalization, which achieved consistent improvement in generalization performance across various natural image and language benchmarks \cite{Author8,Author9,Author10,Author11}. SAM regularizes the sharpness of the loss landscape by simultaneously minimizing the training loss and the loss sharpness. SAM achieves this by employing adversarial perturbations $\epsilon$ to maximize the training loss $L_S(\bold{w}+\epsilon)$, where $\bold{w}$ is the weight vector. It then minimizes the loss of this perturbed objective using an update step from a base optimizer, such as Adam ~\cite{Author12}. To compute $\epsilon$, SAM takes a linear approximation of the loss objective, and uses the gradient $\nabla L_S(\bold{w})$ as the ascent direction for computing $\epsilon = \rho \cdot$norm($\nabla L_S(\bold{w}))$, where $\rho$ is the radius of the maximization region, and norm$(x)=x/||x||_2$ .

However, SAM introduces a significant computational overhead, as the ascent step requires additional forward and backward passes, which doubles the training time. Furthermore, the two-step training process requires tuning additional hyper-parameters, which if not selected correctly, can lead to suboptimal performance or inefficient training. For instance, Wu \etal ~\cite{Author13} and Chen \etal ~\cite{Author14} demonstrate that the perturbation radius $\rho$ should be adjusted beyond the range initially proposed by  Foret \etal ~\cite{Author7} to achieve optimal results. Additionally, while SAM successfully regularizes the sharpness of the loss landscape, it does not address the issue of noisy and high-variance gradients, which can hinder the generalization performance. 

To mitigate these limitations, we propose Gradient Centralized Sharpness Aware Minimization (GCSAM), which integrates Gradient Centralization (GC) ~\cite{Author15} into the ascent step of SAM. GC is a simple yet effective technique that normalizes the gradients by removing their mean, resulting in a more stable and smoother optimization process. Such normalization helps reduce gradient noise and variance, addressing the issue of noisy gradients in SAM. GC not only accelerates convergence but also enhances generalization performance by promoting more consistent and aligned gradient updates ~\cite{Author15, Author16, Author17}. Furthermore, GC reduces the sensitivity to hyperparameters, such as the perturbation radius $\rho$, making the optimization process more robust and efficient. By combining GC with SAM, we not only regularize the sharpness of the loss landscape but also stabilize the training process, ultimately improving the generalization performance across various tasks and datasets. In summary, our main contributions are as follows: 
\begin{itemize}
  \item We propose GCSAM, a novel algorithm that combines Gradient Centralization (GC) with Sharpness-Aware Minimization (SAM). This integration targets both sharpness reduction and gradient stabilization, resulting in improved generalization.  
  \item GCSAM addresses limitations in SAM by reducing computational overhead and mitigating issues of gradient noise. By centralizing gradients, GCSAM provides a more stable optimization process that reduces gradient explosion, enhancing both efficiency and robustness.  
  \item We validate GCSAM’s performance on a range of benchmarks, demonstrating improved generalization in both general and medical imaging datasets. Our results indicate GCSAM’s ability to outperform SAM and baseline optimizers in achieving higher test accuracy with enhanced computational efficiency.
\end{itemize}

%-------------------------------------------------------------------------
\subsection{Related Work}

\textbf{Loss Sharpness and Generalization.} The connection between the geometry of the loss landscape and generalization has been the subject of extensive research effort. Chaudhari \etal ~\cite{Author18} proposed Entropy-SGD, an approach aimed at minimizing the local entropy of the loss landscape to guide the optimization process toward flatter regions, thereby improving generalization. Similarly, Smith and Le ~\cite{Author19} demonstrated that incorporating noise in SGD prevents the optimization from entering sharp valleys. Additionally, Lyu \etal ~\cite{Author20} revealed that gradient descent has an inherent bias toward reducing sharpness, especially in the presence of normalization layers ~\cite{Author21} and weight decay ~\cite{Author22}, which further supports the importance of managing the sharpness of the loss landscape to enhance generalization in DNNs. 

\textbf{Improving SAM.} Several variants of SAM have been proposed to enhance its performance and address its limitations. Adaptive-SAM (ASAM) ~\cite{Author8} modifies SAM by introducing a scaling operator that eliminates sensitivity to model parameter re-scaling, allowing more robust generalization across different models. Kim \etal ~\cite{Author23} refined both SAM and ASAM by leveraging Fisher information geometry, providing a more effective method of minimizing sharpness in the parameter space. Zhuang \etal ~\cite{Author9} highlighted that minimizing the perturbed loss in SAM does not always guarantee a flatter loss landscape, leading to the development of Surrogate-Gap SAM (GSAM), which incorporates a measure akin to the dominant eigenvalue of the Hessian matrix to capture sharpness more accurately. Additionally, Wu \etal observed that SAM's one-step gradient approach may lose effectiveness due to the non-linearity of the loss landscape. To address this, they proposed Curvature Regularized SAM (CR-SAM) ~\cite{Author10}, which integrates a normalized Hessian trace to more precisely measure and regulate the curvature of the loss landscape. Kaddour \etal introduced Weight-Averaged SAM (WASAM) ~\cite{Author24}, which integrates SAM with Stochastic Weight Averaging (SWA) ~\cite{Author25}. Their findings demonstrate that WASAM enhances SAM's generalization performance, particularly in Natural Language Processing (NLP) tasks.

 In addition to improving performance, several algorithms have been proposed to mitigate the computational overhead associated with SAM. Liu \etal introduced LookSAM ~\cite{Author26}, which accelerates SAM by periodically computing the inner gradient ascent at every $k$th iteration instead of at each step, reducing the frequency of expensive updates. Du \etal proposed Efficient SAM (ESAM) ~\cite{Author27}, which minimizes the number of input samples in the second forward and backward passes by utilizing Stochastic Weight Perturbation (SWP) and Sharpness-sensitive Data Selection (SDS) to balance efficiency and sharpness minimization. Furthermore, Becker \etal developed Momentum SAM (MSAM) ~\cite{Author28}, incorporating Nesterov Accelerated Gradient (NAG) ~\cite{Author29} to perturb parameters along the direction of the accumulated momentum vector, which improves both the convergence rate and computational efficiency of SAM.

 \textbf{Gradient Optimization.} Significant research effort has been dedicated to stabilizing and accelerating DNN training through operations on gradients. Gradient clipping ~\cite{Author30,Author31,Author32} was introduced to mitigate the issue of exploding gradients by limiting their magnitude during backpropagation. Qian \etal ~\cite{Author33} explored the use of momentum to accelerate gradient descent optimizers and reduce oscillations, enhancing convergence speed. Further, Riemannian methods ~\cite{Author34} and projected gradient techniques ~\cite{Author35,Author36,Author37} were employed to regulate weight learning by projecting gradients onto subspaces, promoting more stable learning trajectories. Smith \etal ~\cite{Author38} leveraged the gradient norm to derive an implicit regularization term in stochastic gradient descent (SGD), aiding generalization. Additionally, $l_2$ regularization of weight gradients remains one of the most widely adopted strategies to improve the generalization capabilities of DNNs ~\cite{Author39,Author40}. 

\begin{figure*}
  \centering
  \begin{subfigure}{0.4\linewidth}
    \centering
    \includegraphics[width=0.9\linewidth]{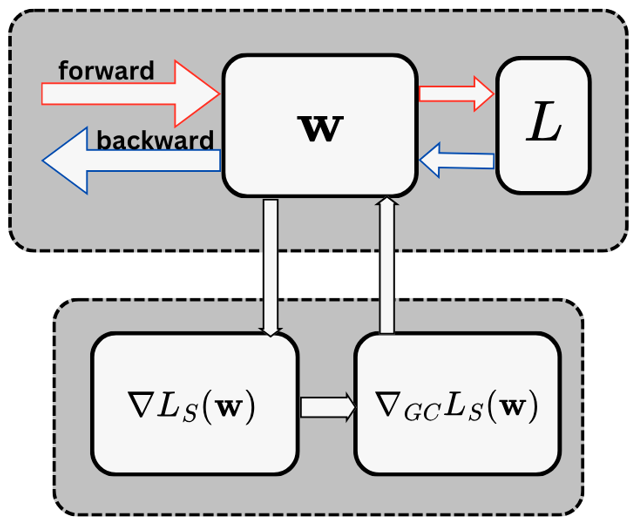}
    \caption{Sketch map for GC}
    \label{fig:gc-a}
  \end{subfigure}
  \hfill
  \begin{subfigure}{0.4\linewidth}
    \centering
    \includegraphics[width=0.9\linewidth]{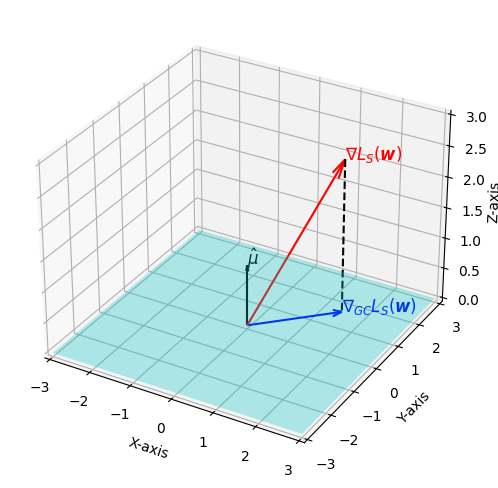}
    \caption{The geometrical interpretation of GC}
    \label{fig:gc-b}
  \end{subfigure}
  \caption{(a) Sketch map for using Gradient Centralization (GC). $L$ is the loss function, $\textbf{w}$ denotes the weight matrix, $\nabla L_S(\bold{w})$ is the gradient of the entire weight matrix, and $\nabla_{GC} L_S(\bold{w})$ is the centralized gradient. (b) Geometrical interpretation of GC. The gradient is projected on a hyperplane $\hat{\mu}$, where the projected gradient is used to update the weight.}
  \label{fig:gc}
\end{figure*}

\section{Method}
\label{sec:method}

In this section, we outline the methodological approaches of three techniques: Sharpness-Aware Minimization (SAM), Gradient Centralization (GC), and our proposed Gradient Centralized Sharpness-Aware Minimization (GCSAM).

%-------------------------------------------------------------------------
\subsection{Sharpness Aware Minimization (SAM)}
While empirical risk minimization algorithms, such as SGD and Adam, effectively reduce the empirical loss $L_S(\bold{w})$ to achieve low training error, addressing the generalization gap $L_D(\bold{w})-L_S(\bold{w})$ remains a challenge in DNN training. Keskar \etal ~\cite{Author41} proposed that there is a connection between the sharpness of minimized empirical loss and the generalization gap, as the large sensitivity of the training function at a sharp minimizer negatively impacts the trained model’s ability to generalize on new data. To formalize this connection, sharpness is defined within an $\epsilon$-ball as: 

\begin{equation}
  \max_{{||\epsilon||}_{p} \le \rho} L_S(\bold{w} + \epsilon) - L_S(\bold{w}),
  \label{eq:important}
\end{equation}
where $\rho$ is the radius of maximization region of an $\ell^p$ ball. From the above definition, sharpness is the difference between the maximum empirical loss in the $\ell^p$ ball and the empirical loss.   

Foret \etal ~\cite{Author7} introduced Sharpness-Aware Minimization (SAM), which integrates sharpness minimization with a PAC-Bayes norm to improve generalization. SAM is designed to enhance generalization by minimizing the sharpness of the loss landscape, encouraging convergence to flat minima. This is achieved by optimizing the following PAC-Bayesian generalization error bound:

\begin{equation}
  L_D(\bold{w}) \le \max_{{||\epsilon||}_{p} \le \rho} L_S(\bold{w} + \epsilon) + h(\frac{||\bold{w}||^2_2}{\rho^2}),
  \label{eq:important}
\end{equation}
where the monotonic nature of $h$ allows substitution with an $\ell^2$ 
weight-decay regularizer. Thus, SAM can be formulated as a minimax optimization problem, aiming to mitigate generalization error while maintaining training stability: 

\begin{equation}
  \min_{\bold{w}} \max_{{||\epsilon||}_{p} \le \rho} L_S(\bold{w} + \epsilon) + \frac{\lambda}{2}(\frac{||\bold{w}||^2_2}{\rho^2}).
  \label{eq:important}
\end{equation}

%-------------------------------------------------------------------------
\subsection{Gradient Centralization (GC)}

Gradient Centralization ~\cite{Author15} is an optimization technique designed to improve both generalization and stability in deep neural networks (DNNs) by centering gradients around a mean of zero during training. In this context, the layer’s loss function is represented as $L_S(\bold{w})$, with $L_S(\bold{w}_i)$ specifically denoting the loss contribution from the $i$-th column of $\bold{w}$. Under the standard DNN setup, the gradient of the entire weight matrix is given by $\nabla L_S(\bold{w})$, and the gradient of the individual weight vector as $\nabla L_S(\bold{w}_i)$. Similarly, when applying Gradient Centralization to a DNN, the gradient of a vector is denoted as $\nabla_{GC} L_S(\bold{w}_i)$ and is defined as: 
\begin{equation}
  \nabla_{GC} L_S(\bold{w}_i) = \nabla L_S(\bold{w}_i) - \frac{1}{n}\sum\limits^{n-1}_{j=0} \nabla L_S(\bold{w}_{ij}).
  \label{eq:important}
\end{equation}
Correspondingly, Gradient Centralization is applied by projecting the gradient, $\nabla L_S(\bold{w})$ onto a subspace orthogonal to the mean gradient direction. This projection is achieved using the matrix $\mathbf{P} = \mathbf{I} - \mathbf{e} \mathbf{e}^\top$, where $\mathbf{e}$ represents a vector of equal components. Mathematically, the centralized gradient is defined as: 

\begin{equation}
  \nabla_{GC} L_S(\bold{w}) = \mathbf{P}\nabla L_S(\bold{w}) = \nabla L_S(\bold{w}) - \mu,
  \label{eq:important}
\end{equation}
where $\mu$ is the mean gradient. From the above equation, the only difference between centralized gradient $\nabla_{GC} L_S(\bold{w})$ and standard gradient $L_S(\bold{w})$ is a deducted mean value from the weight matrix as illustrated in \cref{fig:gc-b}. 

\begin{figure*}
  \centering
  \begin{subfigure}{0.48\linewidth}
    \begin{tabular}{@{}lc}
    \toprule
    \textbf{Algorithm 1:} GCSAM algorithm  \\
    \midrule
    \textbf{Input:} Loss function $L$, Training dataset $S$, Mini-batch \\
    size $b$, Radius of maximization region $\rho$, Step size $\alpha > 0 $, \\
    Initial weight $\bold{w_0}$ \\ 
    \textbf{Output:} Trained weight $\bold{w}$ \\
    \textbf{while} not converged \textbf{do} \\ 
    \quad Sample a mini-batch $B$ of size $b$ from $S$ \\
    \quad Compute gradient $\bold{g} = \nabla L_B(\bold{w}$) of the training loss \\
    \quad Centralize gradient $\bold{g_{GC}} = \bold{g}- \mu$  \\
    \quad Perturb weights for Ascent Step $\epsilon_{GC} = \rho \frac{\bold{g_{GC}}}{||\bold{g_{GC}}||_2}$ \\
    \quad Compute Descent Step $\bold{w} = \bold{w} - \alpha (\bold{g}_{GC}(\bold{w}+\epsilon_{GC}))$ \\
    \textbf{end while} \\
    \textbf{return w} \\
    \bottomrule
  \end{tabular}
    \caption{GCSAM Algorithm}
    \label{tab:gcsam-a}
  \end{subfigure}
  \hfill
  \begin{subfigure}{0.48\linewidth}
    \includegraphics[width=0.9\linewidth]{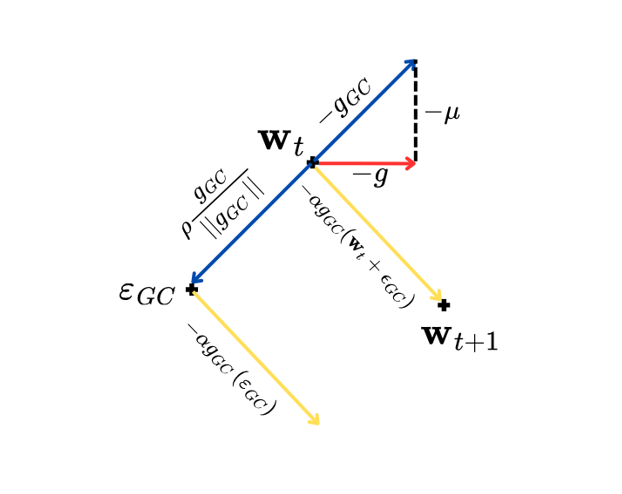}
    \caption{Schematic of the GCSAM parameter update}
    \label{fig:gcsam-b}
  \end{subfigure}
  \caption{(a) GCSAM Algorithm. (b) Schematic of the GCSAM paramter update, where $\textbf{w}_t$ is the current step, $\epsilon_{GC}$ perturbs weights fot the ascent step, and $\textbf{w}_{t+1}$ is the next step.}
  \label{fig:gcsam}
   \end{figure*}
%-------------------------------------------------------------------------
\subsection{Gradient Centralized Sharpness Aware Minimization (GCSAM)}

Motivated by SAM and GC we propose Gradient Centralized Sharpness Aware Minimization (GCSAM), to stabilize the training process and address the noisy gradients in SAM. GCSAM combines these principles by applying gradient centralization into the ascent step of SAM. Specifically, GCSAM minimizes a centralized version of the sharpness objective by applying GC to the gradient used in SAM’s inner maximization problem. This leads to a reformulated objective: 

\begin{equation}
  \min_{\bold{w}} \max_{{||\epsilon_{GC}||}_{p} \le \rho} L_S(\bold{w} + \epsilon_{GC}) + \frac{\lambda}{2}(\frac{||\bold{w}||^2_2}{\rho^2}),
  \label{eq:important}
\end{equation}
where $\epsilon_{GC}$ denotes perturbations aligned with the centralized gradient $\nabla_{GC} L_S(\bold{w})$:

\begin{equation}
  \epsilon_{GC} = \rho \frac{\nabla_{GC} L_S(\bold{w})}{||\nabla_{GC} L_S(\bold{w})||_p},
  \label{eq:important}
\end{equation}

\noindent The final GCSAM algorithm is formulated by integrating a standard numerical optimizer, such as SGD or Adam, as the base optimizer for the GCSAM objective. This combination allows for efficient gradient-based optimization while leveraging the benefits of gradient centralization within the SAM framework. The pseudo-code for the GCSAM algorithm, provided in \cref{tab:gcsam-a}, outlines the proposed approach with SGD as the base optimizer, highlighting the procedural steps involved in each update cycle. Additionally, \cref{fig:gcsam-b} provides a schematic illustration of a single GCSAM parameter update, visually capturing the interaction between gradient centralization and sharpness-aware minimization in the update process. In particular, GCSAM ensures that the magnitude of the centralized gradient is always less than or equal to the original gradient magnitude, leading to tighter sharpness bounds. Details of the derivations and proof are included in Theorem \ref{thm:bigtheorem} in the Appendix.

\section{Experimental Results}

To evaluate the effectiveness of GCSAM, we apply it across a diverse set of tasks, including image classification on CIFAR-10 and medical imaging challenges with Breast Ultrasound (BUS) and COVID-19 datasets. In each case, we assess the impact of GCSAM by benchmarking it against SAM and a standard Adam optimizer, using a variety of deep learning architectures. These include CNN-based models, such as ResNet-50 ~\cite{Author42} and VGG-16~\cite{Author43}, and Vision Transformer architectures like ViT ~\cite{Author44} and Swin Transformer ~\cite{Author45}. Such experimental evaluation allows us to systematically explore the benefits of GCSAM in both general and domain-specific applications. 
%-------------------------------------------------------------------------
\subsection{Image Classification in General Domain}

To confirm the effectiveness of GCSAM in the general domain, we conduct a comparative experiment with Adam, SAM and GCSAM using CIFAR-10 dataset ~\cite{Author46}, containing 60,000 images across 10 classes. We asses the performance of Resnet50 ~\cite{Author42}, VGG-16 ~\cite{Author43}, ViT ~\cite{Author44} and Swin Transformer ~\cite{Author45}. Each model is trained from scratch with a training batch size of 128. We use grid search to identify optimal values for the learning rate ($\alpha$), and perturbation magnitude ($\rho$) for each model. Training continues until each model achieves 100$\%$ accuracy on the training set, with Early Stopping applied to mitigate overfitting, consistent with the experimental settings used by Jiang \etal~\cite{Author3} and Dziugaite \etal~\cite{Author4}.
  
\begin{table}
  \centering
  \begin{tabular}{lclclclc}
    \toprule
    Model & Adam & SAM & GCSAM \\
    \midrule
    Resnet50 & 90.1 & 90.8 & \textbf{91.2} \\ 
    VGG16 & 92.98 & 94.8 & \textbf{95.1} \\ 
    ViT & 83.25 & 84.05 & \textbf{84.08} \\ 
    Swin Transformer & 82.84 & 85.22 & \textbf{85.66} \\ 
    \bottomrule
  \end{tabular}
  \caption{Results for Adam, SAM, and GCSAM on CIFAR-10}
  \label{tab:cifar}
\end{table}

In \cref{tab:cifar}, we report the test accuracies of GCSAM compared to the baseline Adam optimizer and SAM across the four models. The results demonstrate that GCSAM consistently achieves better generalization performance, outperforming both Adam and SAM across all trained architectures.
%-------------------------------------------------------------------------
\subsection{Image Classification in Medical Domain}

Generalization is particularly challenging in the medical domain, where images are collected using a variety of devices, imaging protocols, and across diverse patient populations. These factors introduce unique complexities that impact model performance on unseen data. Prior studies have demonstrated that SAM consistently outperforms traditional optimizers like Adam and SGD across various models due to its ability to achieve flatter minima  ~\cite{Author52, Author53}. To assess the efficacy of GCSAM in tackling these domain-specific challenges, we conduct extensive experiments using two medical datasets: breast ultrasound (BUS) images and COVID-19 chest X-ray images. These datasets are chosen to evaluate the algorithm's robustness under varying imaging conditions and domain shifts. 

\textbf{Breast Ultrasound (BUS) dataset.}  We combined 3,641 breast ultrasound images from ~\cite{Author47} with 2,405 images from the GDPH$\&$SYSUCC dataset~\cite{Author48}, yielding a total of 6,046 breast ultrasound images classified as benign or malignant cases. We partitioned this dataset into an 80$\%$-20$\%$ train-test split. Using an Adam optimizer, SAM and GCSAM, with a learning rate of $10^{-4}$ and a batch size of 16, we trained ResNet50, VGG16, ViT and Swin transformer. Each model was trained until achieving 100$\%$ training accuracy.

\begin{table}[btp]
  \centering
  \begin{tabular}{lclclclc}
    \toprule
    Model & Adam & SAM & GCSAM \\
    \midrule
    Resnet50 & 67.1 & 67.7 & \textbf{70.1} \\ 
    VGG16 & 81.6 & 82.6 & \textbf{83.2} \\ 
    ViT & 70.4 & 70.2 & \textbf{72.1} \\ 
    Swin Transformer & 69.3 & 70.2 & \textbf{70.7} \\ 
    \bottomrule
  \end{tabular}
  \caption{Results for Adam, SAM, and GCSAM on BUS}
  \label{tab:bus}
\end{table}

The results presented in \cref{tab:bus} demonstrate that GCSAM consistently outperforms both the baseline Adam optimizer and SAM across all evaluated models. This performance improvement highlights GCSAM’s capability to enhance generalization in medical imaging tasks. 

\textbf{COVID-19 chest X-ray dataset.} We performed additional experiments utilizing 16,955 COVID-19 chest X-ray images from the COVIDx CXR-4 dataset ~\cite{Author49}, categorized into positive and negative cases. For these experiments, we trained ResNet50, VGG16, Vision Transformer (ViT), and Swin Transformer models with a learning rate of $10^{-4}$ and a batch size of 16. The generalization performance of GCSAM was evaluated against the Adam optimizer, SAM, and several prominent SAM variants: Adaptive SAM (ASAM) ~\cite{Author8}, Surrogate-Gap SAM (GSAM) ~\cite{Author9}, Curvature-Regularized SAM (CRSAM) ~\cite{Author10}, and Momentum SAM (MSAM) ~\cite{Author28}.
To benchmark GCSAM's computational efficiency, we compared it against SAM and its variants, using Adam as the baseline optimizer with a relative speed of $1.0$. This evaluation provided insights into any additional computational overhead introduced by GCSAM while also assessing its effectiveness in improving model generalization.

\begin{table*}
  \centering
  \begin{subtable}{0.49\linewidth}
    \centering
    \begin{tabular}{lclclc}
    \toprule
    Optimizer & Test Accuracy & Speed \\
    \midrule
    Adam & 86.1 & \textbf{1.00} \\ 
    SAM & 89.2 & 2.27\\
    ASAM & 90.2 & 5.60\\
    GSAM & 88.4 & 6.41\\
    CRSAM & 86.6 & 1.37\\
    MSAM & 87.3 & 1.14\\
    GCSAM & \textbf{90.4} & 2.10\\ 
    \bottomrule
    \end{tabular}
    \caption{ResNet50}
    \label{tab:covidresnet50}
  \end{subtable}
  \hfill
  \begin{subtable}{0.49\linewidth}
    \centering
    \begin{tabular}{lclclc}
    \toprule
    Optimizer & Test Accuracy & Speed \\
    \midrule
    Adam & 90.7 & \textbf{1.00} \\ 
    SAM & 91.4 & 1.28\\ 
    ASAM & 91.0 & 5.83\\
    GSAM & 91.5 & 6.67\\
    CRSAM & 90.3 & 1.04\\
    MSAM & 89.3 & 1.02\\
    GCSAM & \textbf{91.8} & 1.07\\ 
    \bottomrule
    \end{tabular}
    \caption{VGG16}
    \label{tab:covidvgg16}
  \end{subtable}
  \begin{subtable}{0.49\linewidth}
    \centering
    \begin{tabular}{lclclc}
    \toprule
    Optimizer & Test Accuracy & Speed \\
    \midrule
    Adam & 81.7 & \textbf{1.00} \\ 
    SAM & 82.2 & 1.32\\
    ASAM & \textbf{82.5} & 3.39\\
    GSAM & 78.1 & 3.60\\
    CRSAM & 80.5 & 1.13\\
    MSAM & 80.3 & 1.05\\
    GCSAM & \textbf{82.5} & 1.18\\ 
    \bottomrule
    \end{tabular}
    \caption{ViT}
    \label{tab:covidvit}
  \end{subtable}
  \begin{subtable}{0.49\linewidth}
    \centering
    \begin{tabular}{lclclc}
    \toprule
    Optimizer & Test Accuracy & Speed \\
    \midrule
    Adam & 86.6 & \textbf{1.00} \\ 
    SAM & 87.5 & 2.04\\
    ASAM & 87.4 & 5.43\\
    GSAM & 84.8 & 5.89\\
    CRSAM & 86.0 & 1.36\\
    MSAM & 84.4 & 1.08\\
    GCSAM & \textbf{87.7} & 1.87\\ 
    \bottomrule
    \end{tabular}
    \caption{Swin Transformer}
    \label{tab:covidswin}
  \end{subtable}
  \caption{Generalization performance and computational cost for Adam, SAM, Adaptive SAM (ASAM), Surrogate-Gap SAM (GSAM), Curvature-Regularized SAM (CRSAM), Momentum SAM (MSAM), and Gradient-Centralized SAM (GCSAM) on COVID-19 dataset.}
  \label{tab:covid}
\end{table*}

In \cref{tab:covid}, we report the test accuracy and computational cost for each optimizer across four models: ResNet50, VGG16, ViT, and Swin Transformer, trained on the COVID-19 CXR-4 dataset. We observe that GCSAM consistently achieves higher test accuracy than the baseline Adam optimizer, SAM and its variants across all models, highlighting its effectiveness in improving generalization performance on medical image classification tasks. In terms of computational cost, GCSAM incurs a moderate overhead compared to Adam, but generally operates more efficiently than SAM, especially on the Vision Transformer (ViT) and Swin Transformer models. For each model, we present the training speed relative to Adam, where Adam’s speed is normalized to 1.0. This setup enables a direct comparison of the computational impact of each optimizer. These results indicate that GCSAM maintains competitive efficiency while enhancing accuracy across different neural network architectures on the COVID-19 dataset.

\section{Discussion}

In this section, we analyze the results presented in \cref{tab:cifar}, \cref{tab:bus}, and \cref{tab:covid}, highlighting the improved generalization and training efficiency offered by the GCSAM optimizer in comparison to SAM and the baseline Adam optimizer.

\textbf{Better Generalization Performance.} Across the CIFAR-10 dataset (\cref{tab:cifar}), GCSAM achieves higher test accuracies than both SAM and Adam on all four tested architectures: ResNet50, VGG16, ViT, and Swin Transformer. This demonstrates that GCSAM’s ability to mitigate gradient explosion in the ascent step of SAM results in more effective training and better model robustness on unseen data. Moreover, GCSAM's generalization improvements are more pronounced in medical imaging tasks, where domain shifts and data variability pose significant challenges to traditional optimization methods. For instance, in breast ultrasound (BUS) images (\cref{tab:bus}) and COVID-19 chest X-rays (\cref{tab:covid}), GCSAM consistently outperforms SAM and Adam, demonstrating its resilience to the inherent variability of medical image datasets. Specifically, for the COVID-19 dataset, GCSAM surpasses all tested SAM variants, further establishing its effectiveness in challenging domains. 

To enhance generalization across all tested models, GCSAM emphasizes three key aspects: sharpness reduction, normalized gradient direction, and suppression of gradient explosion. A core objective of GCSAM is to reduce sharpness in the loss landscape, encouraging convergence in flat regions instead of sharp valleys, which are linked to poor generalization due to the model's sensitivity to small changes in data or input. Additionally, GCSAM modifies the gradients at each optimization step by subtracting the mean of the gradient elements for each layer. This operation serves multiple purposes, such as reducing redundancy in weight updates and promoting a smoother optimization trajectory. As a result, GCSAM prevents undue emphasis on specific neurons or channels, facilitating the model's ability to learn more generalizable features. Furthermore, GCSAM mitigates the risk of gradient explosion, a common issue where gradients grow excessively large, destabilizing parameter updates. In the context of SAM, where gradients are perturbed during the ascent step to explore the loss surface, GCSAM’s stabilizing effect prevents the emergence of large gradients, ensuring more controlled updates. This combined effect enables the model to converge in flatter, more stable regions of the loss surface, avoiding the instability typically caused by the ascent step. In practice, these enhancements lead to improved accuracy on test data across diverse datasets and better performance in various domains, including general and medical imaging tasks.

\textbf{Better Computational Efficieny.} GCSAM offers computational advantages over SAM, as shown in \cref{tab:covid}, with lower relative training times across all models. By reducing gradient spikes, GCSAM accelerates convergence, requiring fewer iterations and thus faster training. This efficiency is particularly beneficial for large-scale models and datasets, where SAM’s ascent step can be computationally expensive. GCSAM’s gradient centralization stabilizes updates, further reducing the computational cost and enabling quicker training by maintaining the model in stable regions of the loss landscape. As a result, GCSAM strikes an effective balance between generalization and computational efficiency.

\begin{figure*}
  \centering
  \begin{subfigure}{0.3\linewidth}
    \centering
    \includegraphics[width=\linewidth]{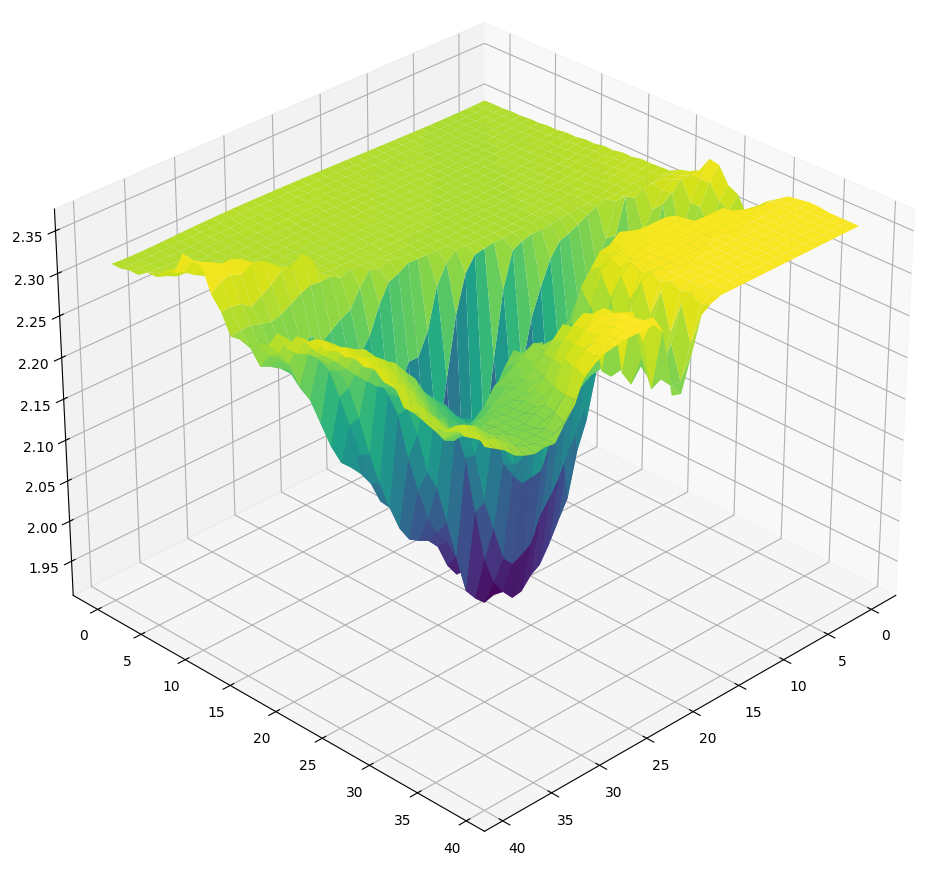}
    \caption{Adam optimized viewing angle 1}
    \label{fig:adam_1}
  \end{subfigure}
  \hfill
  \begin{subfigure}{0.3\linewidth}
    \centering
    \includegraphics[width=\linewidth]{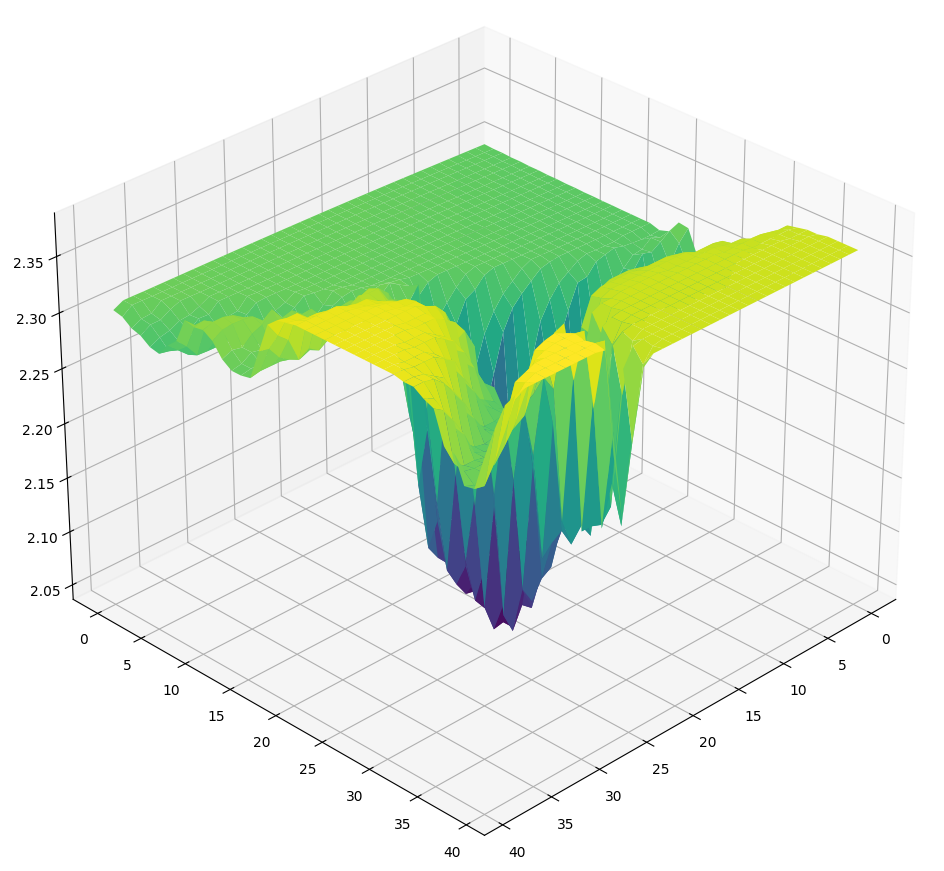}
    \caption{SAM optimized viewing angle 1}
    \label{fig:sam_1}
  \end{subfigure}
  \hfill
  \begin{subfigure}{0.3\linewidth}
    \centering
    \includegraphics[width=\linewidth]{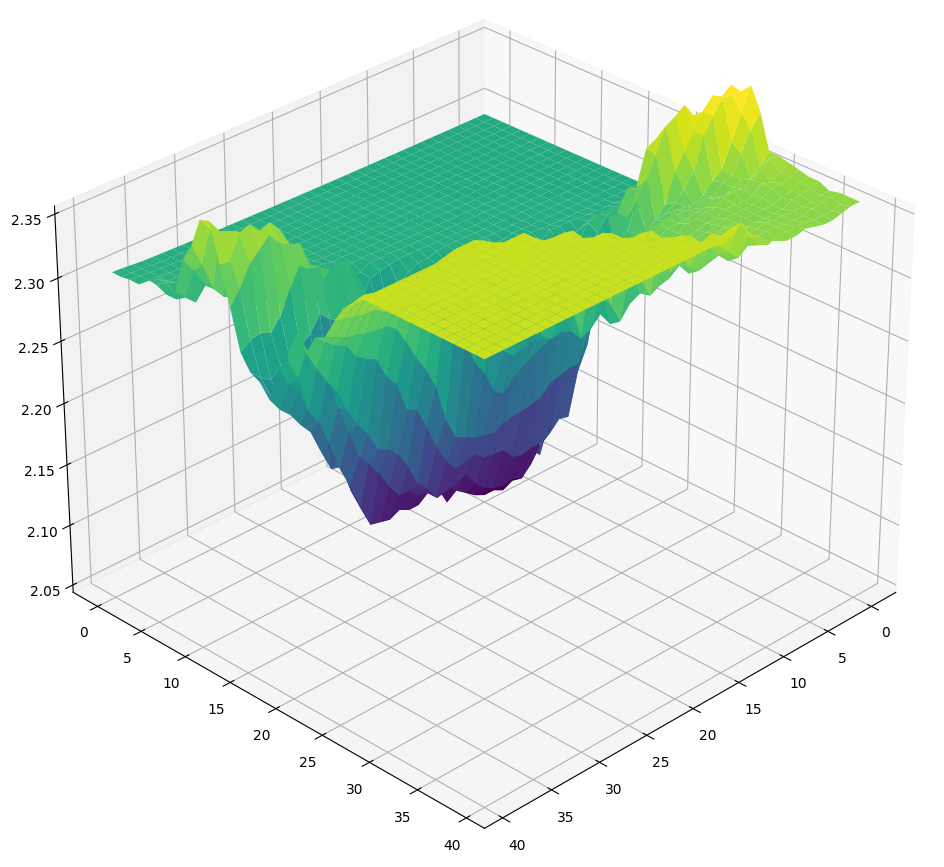}
    \caption{GCSAM optimized viewing angle 1}
    \label{fig:gcsam_1}
  \end{subfigure}
  \hfill
  \begin{subfigure}{0.3\linewidth}
    \centering
    \includegraphics[width=\linewidth]{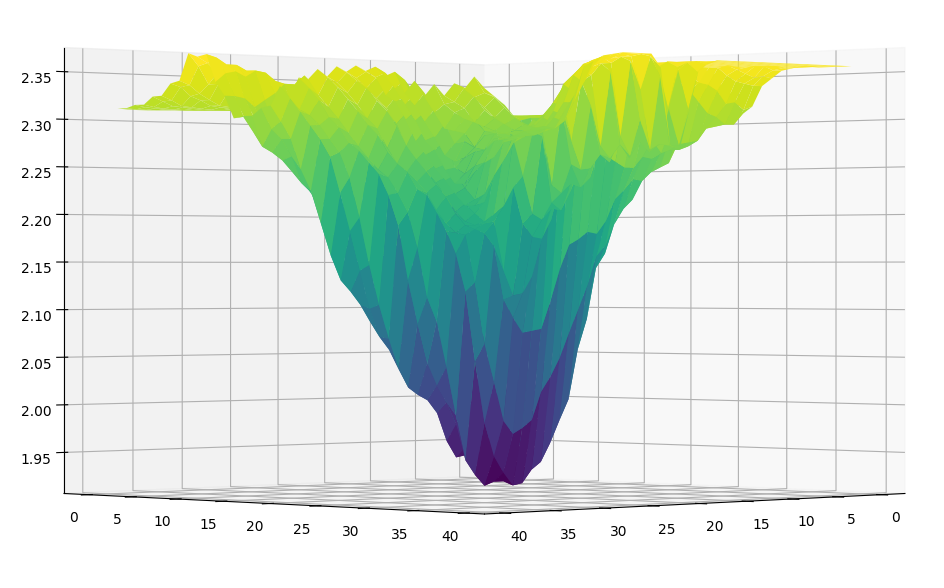}
    \caption{Adam optimized viewing angle 2}
    \label{fig:adam_2}
  \end{subfigure}
  \begin{subfigure}{0.3\linewidth}
    \centering
    \includegraphics[width=\linewidth]{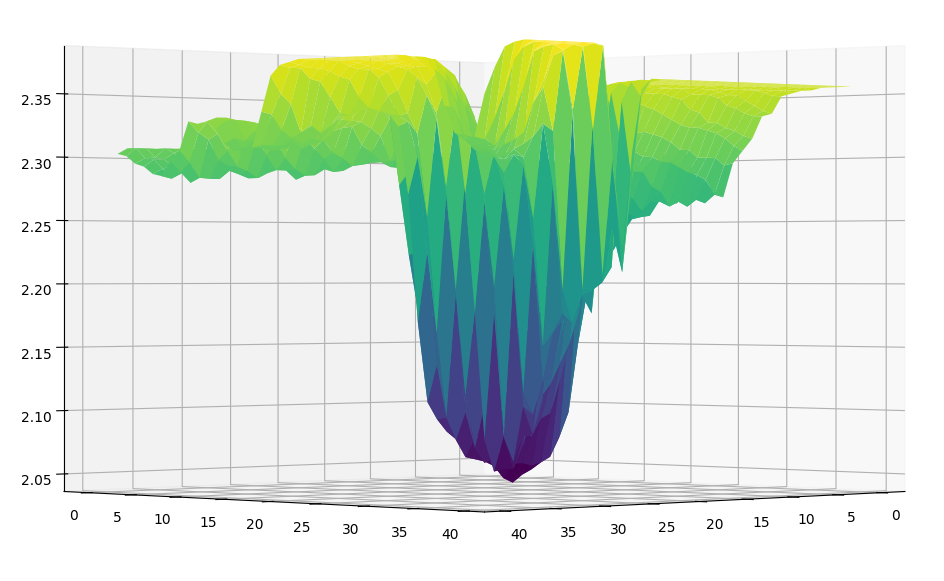}
    \caption{SAM optimized viewing angle 2}
    \label{fig:sam_2}
  \end{subfigure}
  \begin{subfigure}{0.3\linewidth}
    \centering
    \includegraphics[width=\linewidth]{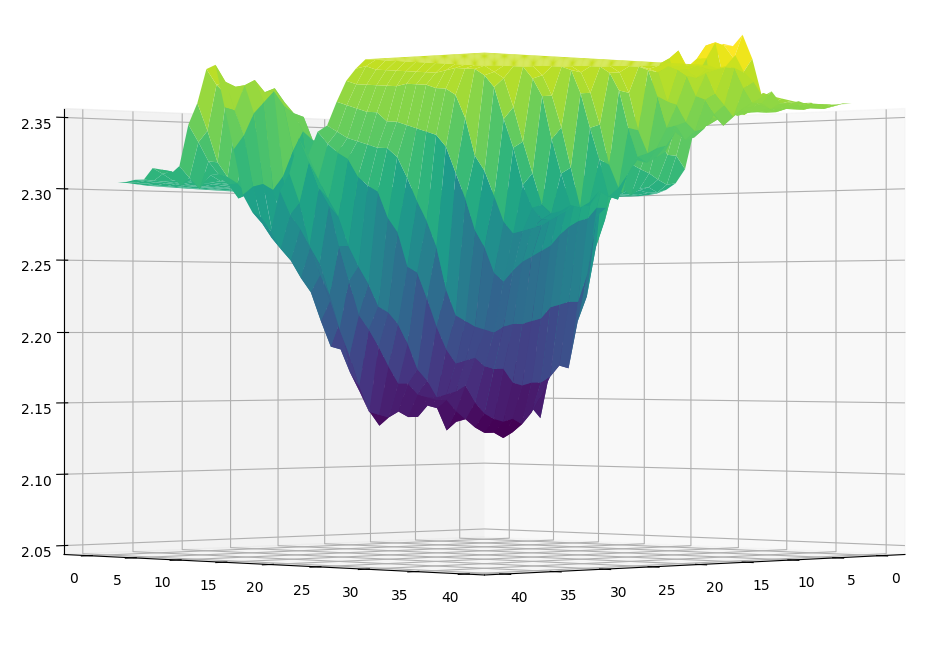}
    \caption{GCSAM optimized viewing angle 2}
    \label{fig:gcsam_2}
  \end{subfigure}
  \caption{Comparison of loss landscapes for Adam, SAM, and GCSAM on the COVID-19 dataset. Adam demonstrates the sharpest minima, reflecting less stable convergence. In contrast, GCSAM achieves the flattest minima, indicative of improved stability and better convergence properties.}
  \label{fig:visualization}
\end{figure*}

\textbf{Visualization of Loss Landscapes.} To analyze the geometry of the loss landscapes obtained with GCSAM, we plotted the loss landscapes of ResNet50 models trained using Adam, SAM, and GCSAM on the COVID-19 dataset. Following the generalization techniques described in ~\cite{Author51}, the loss values were visualized along two orthogonal Gaussian perturbation directions sampled around the local minima. The results in \cref{fig:visualization} reveal that GCSAM produces consistently flatter minima compared to both Adam and SAM.

\section{Conclusion}
In this work, we introduced the Gradient-Centralized Sharpness-Aware Minimization (GCSAM) optimizer, aimed at enhancing model generalization and training stability across diverse applications. Through our comparative analysis with the baseline Adam optimizer and SAM, we observed that GCSAM consistently achieved higher test accuracies in both standard datasets, such as CIFAR-10, and complex medical imaging datasets, including breast ultrasound and COVID-19 chest X-ray images. Our experiments show that GCSAM not only improves performance in terms of test accuracy but also accelerates the training process compared to SAM, attributed to reduced redundancy in weight updates and stabilized parameter changes during the ascent step. These properties make GCSAM a promising approach for applications where data variability and generalization are critical, especially in the medical domain where domain shifts and equipment variability pose significant challenges. Future work could explore incorporating momentum-based strategies within GCSAM to further reduce computational costs, making it even more efficient for large-scale applications. Additionally, adaptive strategies for automatic hyperparameter tuning could enhance its applicability, while extending its use to other domains requiring robust generalization performance.

{
    \small
    \bibliographystyle{ieeenat_fullname}
    \bibliography{main}
}
\appendix
\onecolumn
\section{Appendix}

\subsection{Generalization bound for Gradient Centralized Sharpness Aware Minimization (GCSAM)}
\begin{theorem}
\label{thm:bigtheorem}
Let $\nabla_{GC} L_S(\bold{w}) = \mathbf{P} \nabla L_S(\bold{w})$, where $\mathbf{P} = 1 - \mathbf{e}\mathbf{e}^T $ and $\mathbf{e}$ is a unit vector with $\mathbf{e}^T\mathbf{e} = 1.$ If $L_D(\bold{w}) \le \mathbb{E}_{\epsilon_{i} \sim \mathcal{N}(0, \sigma^2)}[L_D(\bold{w}+\epsilon)] $ for some $\sigma > 0 $, then with probability $1-\delta $ , the GCSAM generalization bound is given by: \\
\[
L_D(\bold{w}) \le \max_{\|\epsilon_{GC}\|_p \leq \rho} L_S(\mathbf{w} + \epsilon_{GC}) + \sqrt{\frac{1}{n-1} (k\log(1 + \frac{||\bold{w}||^2_{2}}{\eta^2\rho^2} (1 + \sqrt{\frac{\log n}{k}})^2)+ 4\log \frac{n}{\delta} + O(1)),}
\] 
where $n = |S| $ is the dataset size, $\rho = \sqrt{k} \sigma (1 + \sqrt{\log n/k})/n $, and $\epsilon_{GC} = \rho \frac{\nabla_{GC} L_S(\bold{w})}{||\nabla_{GC} L_S(\bold{w})||_\rho}$.  

\end{theorem}

\begin{proof} 
Using the proof in SAM by Foret \etal ~\cite{Author7}, from the assumption that adding Gaussian perturbation on the weight space doesn't improve the test error. The following PAC-Bayesian generalization bound holds under the perturbations: \\
\[
\mathbb{E}_{\epsilon_i \sim \mathcal{N}(0, \sigma^2)}[L_D(\bold{w} + \epsilon)] \leq \mathbb{E}_{\epsilon_i \sim \mathcal{N}(0, \sigma^2)}[L_S(\bold{w} + \epsilon)] + \sqrt{\frac{1}{n-1} (k\log(1 + \frac{||\bold{w}||^2_{2}}{\eta^2\rho^2} (1 + \sqrt{\frac{\log n}{k}})^2)+ 4\log \frac{n}{\delta} + O(1)),}
\]
which can be bounded as: \\
\[
L_D(\bold{w}) \le \max_{{||\epsilon||}_{p} \le \rho} L_S(\bold{w} + \epsilon_{GC}) + \sqrt{\frac{1}{n-1} (k\log(1 + \frac{||\bold{w}||^2_{2}}{\eta^2\rho^2} (1 + \sqrt{\frac{\log n}{k}})^2)+ 4\log \frac{n}{\delta} + O(1))} . \\
\]
For the centralized gradient, the magnitude is:
\[
\| \nabla_{GC} L_S(\bold{w}) \|_2^2 = (\nabla_{GC} L_S(\bold{w}))^\top \nabla_{GC} L_S(\bold{w})
\]
Expanding this: \\
\[
\| \nabla_{GC} L_S(\bold{w}) \|_2^2 = (\mathbf{P} \nabla L_S(\bold{w}))^\top (\mathbf{P} \nabla L_S(\bold{w}))
\]
Using the symmetry of $\mathbf{P}$: \\
\[
\| \nabla_{GC} L_S(\bold{w}) \|_2^2 = \nabla L_S(\bold{w})^\top \mathbf{P}^\top \mathbf{P} \nabla L_S(\bold{w}) =  \nabla L_S(\bold{w})^\top \mathbf{P} \nabla L_S(\bold{w})
\]
Expanding $\mathbf{P} = \mathbf{I} - \mathbf{e} \mathbf{e}^\top$: \\
\[
\| \nabla_{GC} L_S(\bold{w}) \|_2^2 = \nabla L_S(\bold{w})^\top (\mathbf{I} - \mathbf{e} \mathbf{e}^\top) \nabla L_S(\bold{w})
\]
\[
= \nabla L_S(\bold{w})^\top \nabla L_S(\bold{w}) - \nabla L_S(\bold{w})^\top \mathbf{e} \mathbf{e}^\top \nabla L_S(\bold{w})
\]
\[
=\|   \nabla L_S(\mathbf{w}) \|_2^2 - \| \mathbf{e}^\top \nabla L_S(\mathbf{w}) \|_2^2,
\]
where $\| \mathbf{e}^\top \nabla L_S(\mathbf{w}) \|_2^2$ is non-negative, so we get that: \\ 
\[
\| \nabla_{GC} L_S(\mathbf{w}) \|_2^2 \leq \| \nabla L_S(\mathbf{w}) \|_2^2.
\]
The inequality demonstrates that the centralized gradient magnitude is always less than or equal to the original gradient magnitude. Using the reduced gradient magnitude, the perturbation $\epsilon_{GC}$ in GCSAM leads to tighter sharpness bounds ($\max_{\|\epsilon\|_p \leq \rho} L_S(\mathbf{w} + \epsilon) \longrightarrow \max_{\|\epsilon_{GC}\|_p \leq \rho} L_S(\mathbf{w} + \epsilon_{GC})$). Using this information we get that: \\ 
\[
L_D(\bold{w}) \le \max_{\|\epsilon_{GC}\|_p \leq \rho} L_S(\mathbf{w} + \epsilon_{GC}) + \sqrt{\frac{1}{n-1} (k\log(1 + \frac{||\bold{w}||^2_{2}}{\eta^2\rho^2} (1 + \sqrt{\frac{\log n}{k}})^2)+ 4\log \frac{n}{\delta} + O(1))}
\]

\end{proof}

\end{document}